\documentclass[runningheads]{llncs}
\usepackage[T1]{fontenc}
\usepackage{graphicx}
\usepackage{amsmath} 
\usepackage{multirow}
\usepackage{booktabs}
\usepackage{amssymb}
\usepackage{xcolor}

\begin{document}
\title{Modeling Nonlinear Feature Interactions with Product-Unit Residual Networks}
\titlerunning{Product-Unit Residual Network}
%
%
\author{Ziyuan Li\inst{1,2} \and
Uwe Jaekel\inst{1} \and
Babette Dellen\inst{1}}
\authorrunning{Li et al.}
%
\institute{Department of Mathematics, Informatics and Technology, University of Applied Sciences Koblenz, Joseph-Rovan-Allee 2, 53424 Remagen, Germany\\ \email{\{jaekel, dellen\}@hs-koblenz.de} \and
Technical University of Munich, Munich, Germany\\ \email{ziyuan.li@tum.de}}
\maketitle              
\begin{abstract}
Understanding nonlinear feature interactions is crucial in science and engineering, yet standard multilayer perceptrons (MLPs) often capture such interactions only implicitly, leading to entangled representations that can impair robustness and interpretability. We investigate product-unit residual networks (PURe) that integrate multiplicative product units with residual connections to explicitly model cross-feature couplings while stabilizing optimization. We conduct a systematic evaluation on an interaction-driven synthetic benchmark and two real-world datasets, assessing predictive accuracy, robustness to Gaussian feature noise, and performance under limited training data, and we compare real- and complex-valued variants under a matched parameter budget. Beyond accuracy, SHapley Additive exPlanations (SHAP)-based interaction analyses show that PURe learns more concentrated and structurally coherent interaction patterns than MLP baselines. Overall, PURe achieves competitive or improved performance, better robustness and sample efficiency in low-data regimes, and enhanced interaction-level interpretability.
\end{abstract}
\section{Introduction}
Modeling nonlinear relationships between input variables is a fundamental challenge occurring in scientific and engineering regression tasks, including physical system modeling, materials science, and socio-economic prediction problems. Often the target variable is not determined solely by individual features, but by nonlinear interactions among multiple factors. Accurately capturing these nonlinear feature interactions is therefore essential for achieving reliable predictions and meaningful scientific insights.

Multilayer perceptron networks (MLP) have become a standard tool for nonlinear regression due to their universal approximation capability and flexible function representation \cite{cybenko1989approximation,rumelhart1986learning}. However, their underlying additive composition often leads to implicit and entangled representations of feature interactions, which can limit generalization performance, particularly in data-scarce or noisy regimes \cite{poggio2017theory,zhang2017understanding}. Moreover, internal representations learned by conventional neural networks are typically difficult to interpret, making it challenging to assess whether meaningful interaction patterns have been captured or whether the model has learned spurious correlations \cite{molnar2020interpretable,lundberg2017unified}. Although recent studies have shown that neural networks can implicitly encode statistical feature interactions, such interactions are often entangled and difficult to disentangle without explicit structural constraints \cite{tsang2017detecting}.

To address these limitations, multiplicative neural architectures based on product units have been proposed as an alternative modeling paradigm \cite{li2025advancing,li2025deep,dellen2024predicting,li2024data,dellen2019function,leerink1995learning,durbin1989product}. By explicitly encoding multiplicative relationships, product-unit networks provide a natural inductive bias for modeling nonlinear feature interactions that commonly arise in scientific data. Despite their theoretical expressiveness, product-unit models have to date received limited attention, presumably due to optimization difficulties, sensitivity to initialization, and training instability \cite{leerink1995learning,dellen2019function}, which are expected to be especially problematic in deeper architectures.

Residual connections have been shown to substantially improve the optimization and stability of deep neural networks by facilitating gradient flow and reducing vanishing gradient effects \cite{he2016deep}. Residual architectures have been successfully applied to a wide range of learning problems, including regression tasks with complex nonlinear structures \cite{veit2016residual,allen2019convergence}. Recently, the use of residual connections in conjunction with multiplicative product-unit networks has been proposed \cite{li2025deep,li2025advancing}. Although the results have been encouraging, it is unclear how residual connections influence predictive performance, generalization behavior, and interpretability when targeting explicit interaction modeling.

In this work, we study product-unit residual networks (PURe), a class of neural architectures that integrate multiplicative product units with residual connections to explicitly model nonlinear feature interactions while maintaining stable optimization, and conduct a systematic empirical evaluation on both synthetic and real-world regression datasets, analyzing predictive accuracy, robustness to noise, and performance under limited training data. Furthermore, an interaction-based analysis is employed to assess whether the proposed architecture captures meaningful feature dependencies with improved structural clarity compared to standard MLP baselines. Our results demonstrate that PURe achieves competitive performance, enhanced generalization, and improved interpretability, making it a promising modeling framework for computational-science applications governed by complex nonlinear relationships.

\section{Related Work}
Product-unit neural networks were introduced to explicitly encode multiplicative interactions between input features \cite{durbin1989product}. Given an input vector $\mathbf{x} = (x_1, \dots, x_d)$, a classical product unit computes
\begin{equation}
y = \prod_{i=1}^{d} x_i^{w_i},
\end{equation}
where $w_i$ denotes a learnable exponent associated with the $i$-th input. This formulation enables compact representations of high-order feature interactions that may otherwise require substantially deeper additive networks.

For numerical stability, product units are commonly implemented using a logarithmic reparameterization \cite{dellen2019function}:
\begin{equation}
y = \exp\left( \sum_{i=1}^{d} w_i \log x_i \right).
\end{equation}
This log-linear-exponential form highlights the multiplicative inductive bias of product units and distinguishes them structurally from additive neural representations. Despite their expressive advantages, previous studies have reported optimization difficulties and sensitivity to initialization, which presumably have limited the use of product-unit networks in deeper architectures so far \cite{dellen2019function}.

Product-unit formulations are inherently restricted to positive-valued inputs due to the logarithmic transformation involved in their implementation. To accommodate signed inputs and to enable joint modeling of magnitude and phase information, product-unit representations have been extended to the complex domain \cite{dellen2024predicting,li2025advancing}. For a complex-valued input $\mathbf{z} \in \mathbb{C}^d$, a complex product unit can be defined as
\begin{equation}
y = \exp\left( \sum_{i=1}^{d} w_i \log (z_i) \right), \quad w_i \in \mathbb{C},
\end{equation}
where the complex logarithm is given by $\log(z) = \log|z| + \mathrm{i}\arg(z)$. This formulation preserves the multiplicative interaction structure while enabling joint modeling of amplitude and phase, which is difficult to achieve using real-valued additive networks. Existing work on complex-valued product units has primarily focused on representational aspects, with limited investigation into their integration with modern deep architectural components.

\section{Method}
\subsection{Product-Unit Residual Block}
Let $\mathbf{h}\in\mathbb{R}^{d}$ denote the input of a residual block, where $d$ is the hidden dimension.
Our real-valued product-unit residual block is implemented as a three-stage transformation followed by an identity skip connection
\begin{equation}
\mathbf{h}_{\text{out}} = \mathbf{h} + \mathrm{FC}_2\!\left(\mathrm{PU}\!\left(\mathrm{FC}_1(\mathbf{h})\right)\right).
\label{eq:pure_block_real}
\end{equation}
Here, $\mathrm{FC}_1$ and $\mathrm{FC}_2$ are affine mappings (linear layers with bias) from $\mathbb{R}^{d}$ to $\mathbb{R}^{d}$.

Given an intermediate vector $\mathbf{a}=\mathrm{FC}_1(\mathbf{h})\in\mathbb{R}^{d}$, the product-unit mapping is computed using a positivity-enforced preprocessing with 
\begin{equation}
\mathbf{s} = \mathrm{ReLU}(\mathbf{a}) + \mathbf{1},
\qquad \mathrm{ReLU}(\cdot)=\max(\cdot,0),
\label{eq:relu_plus_one}
\end{equation}
where $\mathbf{1}\in\mathbb{R}^d$ denotes the all-ones vector, followed by a log-linear-exponential computation
\begin{equation}
\mathrm{PU}(\mathbf{a}) = \exp\!\left( W_{\mathrm{pu}} \log(\mathbf{s}) \right),
\label{eq:pu_real}
\end{equation}
where $W_{\mathrm{pu}}\in\mathbb{R}^{d\times d}$ is a learnable weight matrix (without bias), and $\log(\cdot)$ and $\exp(\cdot)$ are applied element-wise. We adopt $\mathbf{s}=\mathrm{ReLU}(\mathbf{a})+\mathbf{1}$ to ensure $\mathbf{s}>\mathbf{0}$, so that $\log(\mathbf{s})$ is well-defined and zero-valued factors are avoided. Moreover, adding $1$ provides a strictly positive multiplicative gate: when $a_k\le 0$, $s_k=1$ acts as a neutral element (since $\log 1=0$), while for $a_k>0$ it produces a positive scaling factor, enabling the PU to modulate interactions without sign ambiguities.

\subsection{Overall Real-Valued Network}
Given an input $\mathbf{x}\in\mathbb{R}^{p}$, the network first maps it to the hidden space and applies an element-wise rectified linear unit:
\begin{equation}
\mathbf{h}^{(0)}=\mathrm{ReLU}\!\left(\mathrm{FC}_{\text{in}}(\mathbf{x})\right).
\end{equation}
Then, two residual blocks are applied sequentially, namely 
\begin{equation}
\mathbf{h}^{(1)}=\mathcal{B}\!\left(\mathbf{h}^{(0)}\right),\qquad
\mathbf{h}^{(2)}=\mathcal{B}_0\!\left(\mathbf{h}^{(1)}\right),
\end{equation}
where $\mathcal{B}$ denotes the block in Eq.~\eqref{eq:pure_block_real}, and $\mathcal{B}_0$ denotes the same block with zero initialization for stability (see implementation details).
Finally, the prediction is produced by a linear output layer with
\begin{equation}
\hat{y} = \mathrm{FC}_{\text{out}}\!\left(\mathbf{h}^{(2)}\right).
\end{equation}

\subsection{Complex-Valued Product-Unit Residual Block}
To accommodate signed inputs and enable joint modeling of magnitude and phase information, we extend the block to the complex domain.
Let $\mathbf{z}\in\mathbb{C}^{d}$ be the input of a complex-valued block.
The block structure mirrors the real-valued case, i.e., 
\begin{equation}
\mathbf{z}_{\text{out}} = \mathbf{z} + \mathrm{FC}_2^{\mathbb{C}}\!\left(\mathrm{PU}_{\mathbb{C}}\!\left(\mathrm{FC}_1^{\mathbb{C}}(\mathbf{z})\right)\right).
\label{eq:pure_block_c}
\end{equation}

Given $\mathbf{a}=\mathrm{FC}_1^{\mathbb{C}}(\mathbf{z})$, we apply a numerically-stable preprocessing by separately rectifying the real- and imaginary parts and adding a constant offset, to obtain
\begin{equation}
\tilde{\mathbf{a}} = \big(\mathrm{ReLU}(\Re(\mathbf{a})) + c\big) + \mathrm{i}\big(\mathrm{ReLU}(\Im(\mathbf{a})) + c\big),
\qquad c=\sqrt{0.5},
\label{eq:complex_safe}
\end{equation}
and compute
\begin{equation}
\mathrm{PU}_{\mathbb{C}}(\mathbf{a}) = \exp\!\left( W_{\mathrm{pu}}^{\mathbb{C}} \log(\tilde{\mathbf{a}}) \right),
\label{eq:pu_complex}
\end{equation}
where $W_{\mathrm{pu}}^{\mathbb{C}}\in\mathbb{C}^{d\times d}$ is a complex-valued weight matrix (without bias), and $\log(\cdot)$ and $\exp(\cdot)$ denote the complex logarithm and exponential applied element-wise.

\section{Experiments}

\subsection{Experimental Setup}
We evaluate the proposed PURe on three regression benchmarks of increasing complexity, covering synthetic interaction-driven data, small-scale real-world engineering data, and large-scale socio-economic data.

\emph{Friedman~1} is a synthetic benchmark designed to explicitly assess a model's ability to capture nonlinear feature interactions \cite{friedman1991multivariate}. It consists of $10$ independent input features, where the target variable is generated by known nonlinear interactions among a subset of features,
\begin{equation}
y = 10\sin(\pi x_0 x_1) + 20(x_2 - 0.5)^2 + 10x_3 + 5x_4 + \varepsilon,
\end{equation}
while the remaining features act as noise. The dataset contains $10{,}000$ samples and is split into $70\%$ training, $15\%$ validation, and $15\%$ test sets.

\emph{Concrete Compressive Strength} is a real-world engineering dataset involving nonlinear relationships between material composition and compressive strength \cite{yeh1998modeling}. It includes $8$ physical and chemical features and $1{,}030$ samples, split into $60\%$ training, $20\%$ validation, and $20\%$ test sets.

\emph{California Housing} is a large-scale real-world regression benchmark based on U.S.\ census data, featuring heterogeneous socio-economic and geographic factors \cite{pace1997sparse}. It contains $8$ input features and $20{,}640$ samples, split into $80\%$ training, $10\%$ validation, and $10\%$ test sets.

We evaluated real-valued and complex-valued variants of both MLP and PURe under identical experimental settings. For complex-valued models, all layers operate in the complex domain, and the final prediction is obtained by applying a magnitude operation to the complex-valued output in order to produce a real-valued regression target. This is appropriate here because all three benchmarks have non-negative targets. To ensure a fair comparison in terms of model capacity, we account for the fact that each complex-valued parameter corresponds to two real-valued degrees of freedom. Accordingly, real-valued networks use a hidden dimension of $128$, while complex-valued networks use a hidden dimension of $64$, resulting in comparable total numbers of trainable parameters across real and complex architectures.

All models are trained using the mean squared error loss, i.e., 
\begin{equation}
\mathcal{L} = \frac{1}{N}\sum_{i=1}^{N}\left(y_i - \hat{y}_i\right)^2,
\end{equation}
where $y_i$ and $\hat{y}_i$ denote the ground-truth target and model prediction, respectively \cite{bishop2006pattern}.

The models are trained for $100$ epochs using the Adam optimizer with a learning rate of $10^{-3}$ and a batch size of $128$. Unless stated otherwise, all layers are initialized using Kaiming initialization \cite{he2015delving}. For complex-valued layers, we apply Kaiming initialization independently to the real and imaginary parts. For the PURe, the second residual block is initialized to zero to improve early-stage training stability, while the remaining layers follow standard initialization.

Each experimental configuration is repeated over five independent runs with different random seeds. Results are reported as the mean and sample standard deviation across runs. For evaluation on the test set, the model checkpoint achieving the lowest validation loss is selected.

For interpretability analyses, including the SHapley Additive exPlanations (SHAP)-based feature interaction studies, models are trained once using a fixed random seed. These analyses are intended to qualitatively examine the structure and stability of learned feature interactions rather than to provide statistically aggregated performance estimates.

To assess robustness, predictive performance is evaluated under both clean conditions and Gaussian noise corruption, where zero-mean noise with standard deviation $\sigma=0.05$ is added to the input features at test time.

\subsection{Benchmark Performance and Robustness}
\begin{figure}[htbp]
    \centering
    \includegraphics[width=0.78\linewidth]{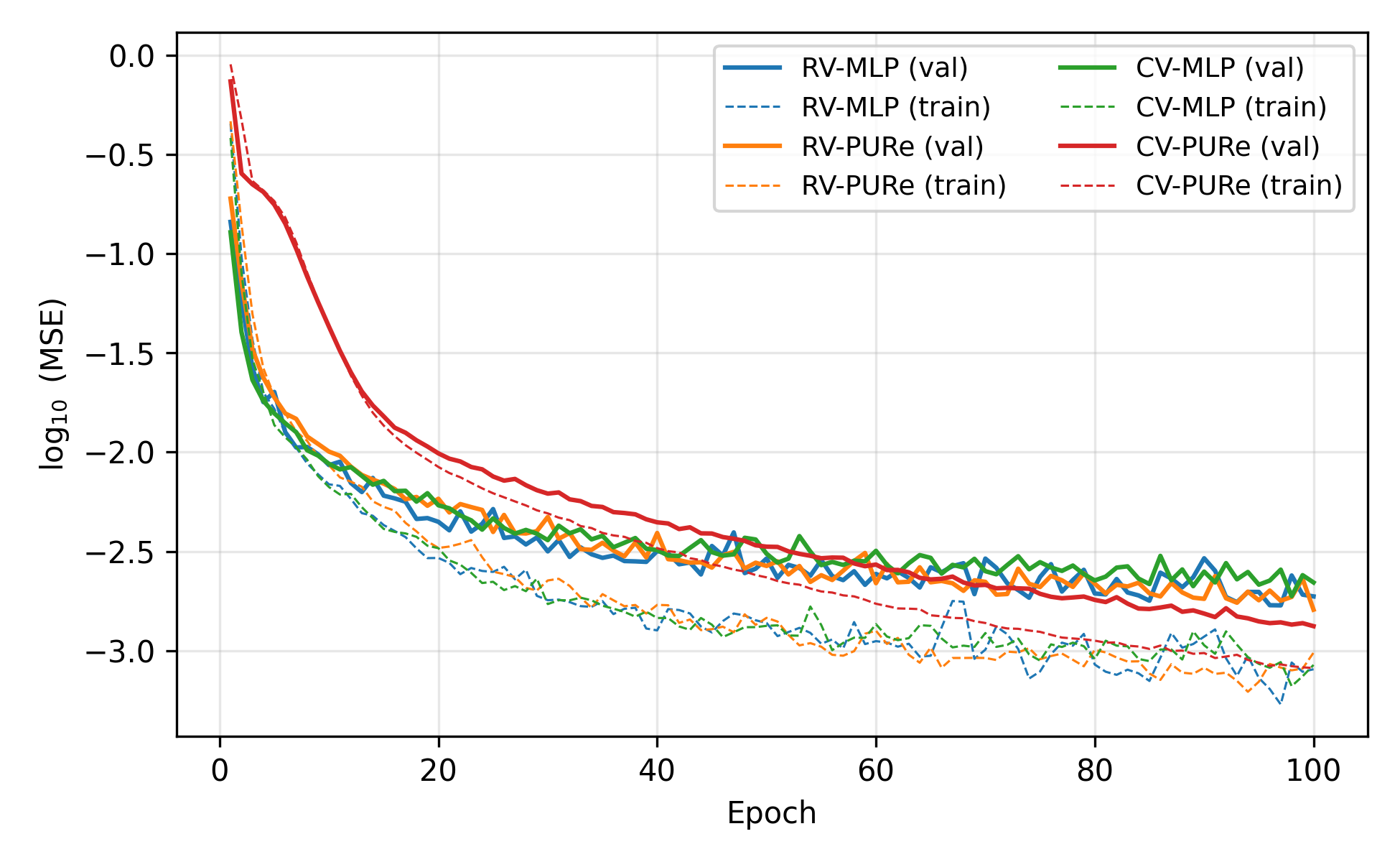}
    \includegraphics[width=0.78\linewidth]{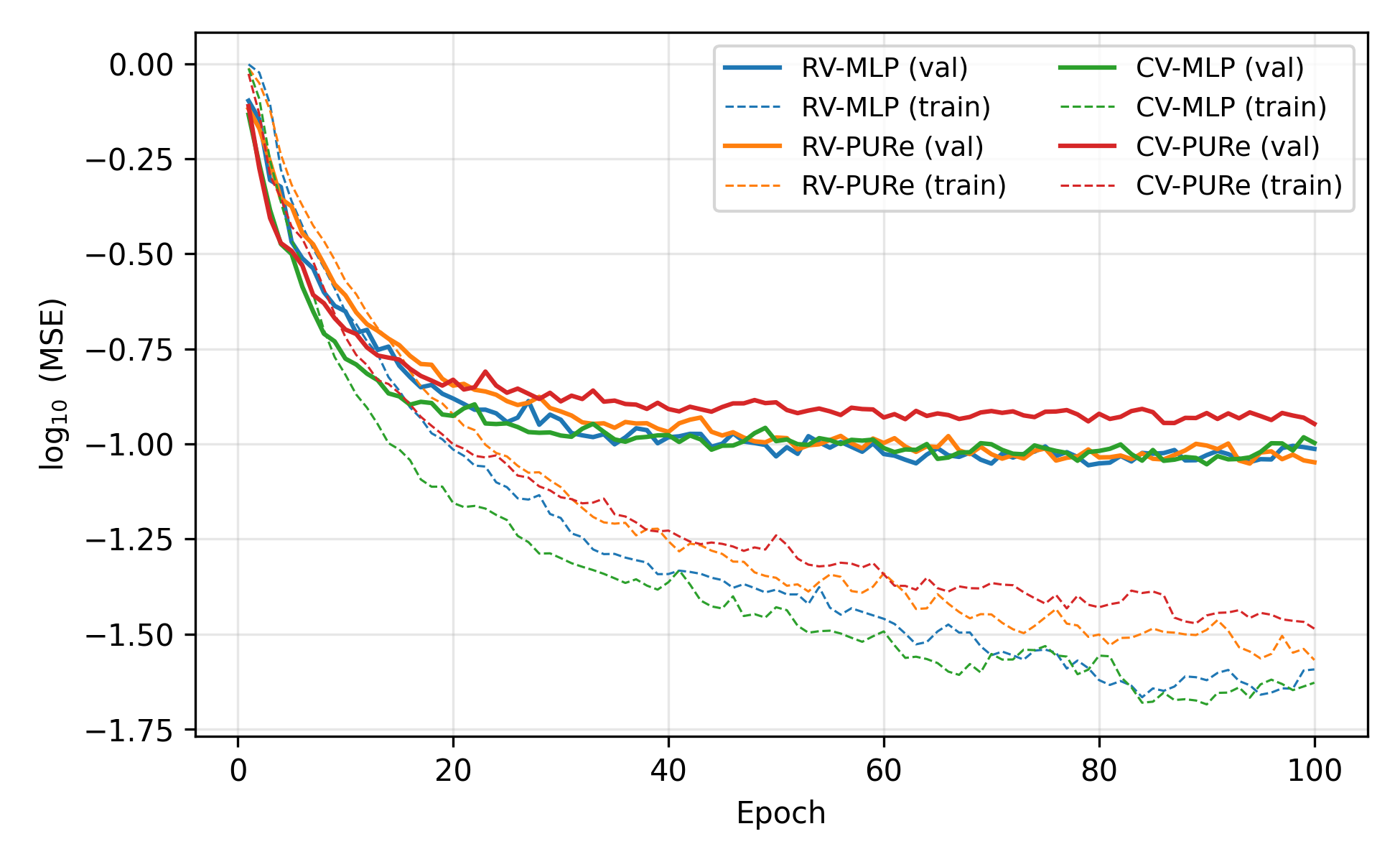}
    \includegraphics[width=0.78\linewidth]{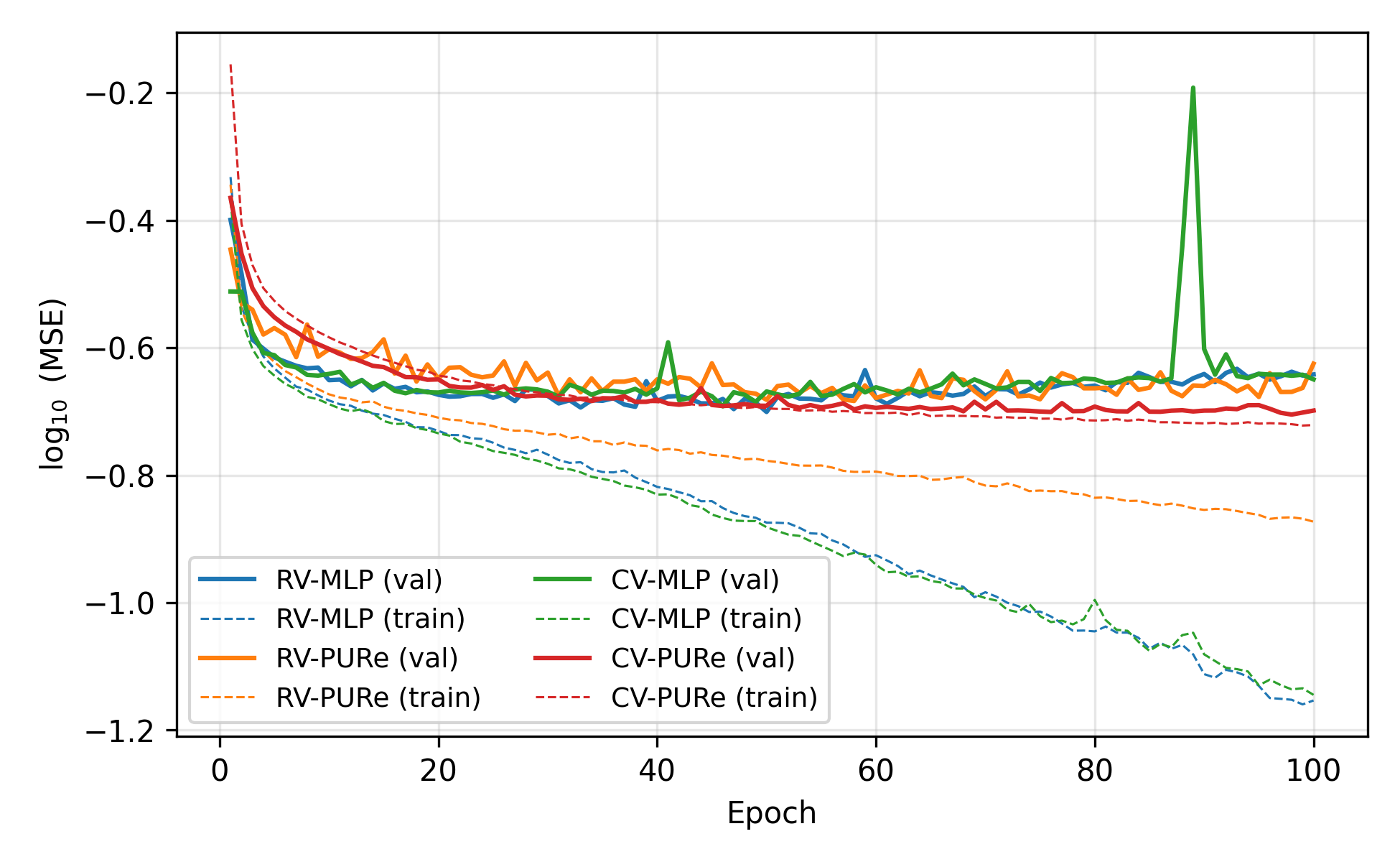}
    \caption{Training and validation loss curves (log10 MSE) on the three benchmark datasets: Friedman 1 (top), Concrete Compressive Strength (middle), and California Housing (bottom).}
    \label{fig:loss_curves}
\end{figure}

We evaluate the proposed PURe with respect to predictive accuracy, robustness, and training dynamics. Quantitative test set results are summarized in Table~\ref{tab:results}, while training and validation loss trajectories are presented in Fig.~\ref{fig:loss_curves}.
\begin{table}[h]
\centering
\setlength{\tabcolsep}{2.6pt}
\renewcommand{\arraystretch}{1.15}
\caption{Test mean squared error (MSE; mean $\pm$ sample standard deviation over five runs).
We report results under clean inputs ($\sigma=0.00$) and Gaussian input noise ($\sigma=0.05$).
The best result for each dataset and noise level is highlighted in bold.}
\label{tab:results}
\begin{tabular}{lccccc}
\toprule
DS & $\sigma$ &
RV-MLP & RV-PURe & CV-MLP & CV-PURe \\\midrule

\multirow{2}{*}{Fried1} & 0.00 & 0.0020 $\pm$ 0.0004 & 0.0016 $\pm$ \textbf{0.0001} & 0.0019 $\pm$ 0.0002 & \textbf{0.0013} $\pm$ 0.0002 \\
& 0.05 & 0.0066 $\pm$ 0.0004 & 0.0062 $\pm$ \textbf{0.0002} & 0.0066 $\pm$ \textbf{0.0002} & \textbf{0.0058} $\pm$ \textbf{0.0002} \\

\midrule
\multirow{2}{*}{Conc.} & 0.00 & 0.1118 $\pm$ 0.0041 & 0.1095 $\pm$ \textbf{0.0015} & 0.1130 $\pm$ 0.0090 & \textbf{0.1092} $\pm$ 0.0035 \\
& 0.05 & 0.1275 $\pm$ 0.0068 & 0.1265 $\pm$ \textbf{0.0028} & 0.1304 $\pm$ 0.0083 & \textbf{0.1258} $\pm$ 0.0041 \\

\midrule
\multirow{2}{*}{Calif.} & 0.00 & 0.2117 $\pm$ 0.0108 & 0.1968 $\pm$ 0.0075 & 0.1926 $\pm$ 0.0139 & \textbf{0.1869} $\pm$ \textbf{0.0043} \\
& 0.05 & 0.2918 $\pm$ 0.0113 & 0.2777 $\pm$ 0.0095 & 0.2609 $\pm$ 0.0129 & \textbf{0.2484} $\pm$ \textbf{0.0053} \\

\bottomrule
\end{tabular}
\end{table}

On the Friedman 1 dataset, CV-PURe eventually achieves the lowest test MSE as shown in Table~\ref{tab:results}. However, the loss trajectories in Fig.~\ref{fig:loss_curves} (top) reveal that while the MLP baselines converge more rapidly in the initial 20 epochs, the CV-PURe model (red solid line) exhibits a more sustained and stable descent, eventually surpassing all other models after approximately 50 epochs. This suggests that while the explicit multiplicative structure may require more iterations to resolve initially, it ultimately yields a more precise fit to the underlying interaction-driven data.

On the Concrete dataset, although CV-PURe achieves the best average test performance (Table~\ref{tab:results}), the corresponding loss curves in Fig.~\ref{fig:loss_curves} (middle) show that it plateaus at a slightly higher validation MSE compared to the RV-based models during this specific run. Notably, however, the CV-PURe trajectory is significantly smoother, avoiding the erratic oscillations seen in the CV-MLP and RV-MLP curves. This stability highlights the regularization benefit of the CV-PURe architecture, even in regimes where the performance gains are marginal.

On the California Housing dataset, the validation trajectories largely overlap, confirming that all models achieve comparable accuracy. Nevertheless, the CV-MLP model (green solid line) exhibits a prominent instability spike near the 90th epoch, whereas the PURe variants remain remarkably stable. This reinforces the conclusion that explicit multiplicative modeling, particularly when combined with complex-valued weights, enhances the robustness and training stability of the network, especially when dealing with the high-variance geographic dependencies inherent in this dataset.

Across the benchmark datasets, the product-unit models, particularly CV-PURe, demonstrate superior robustness to Gaussian input noise. As detailed in Table~\ref{tab:results}, CV-PURe consistently maintains the lowest mean test MSE under both clean and noisy ($\sigma=0.05$) conditions, showing significantly less performance degradation than the MLP baselines. While the more expressive complex-valued parameterization in CV-PURe introduces slightly higher training variability (standard deviation) in certain cases like Friedman 1 and Concrete, it consistently yields the highest predictive accuracy. Ultimately, CV-PURe offers a superior trade-off between expressiveness and noise resilience, effectively balancing model stability with the ability to capture complex nonlinear dependencies.

\subsection{Interpretability and Feature Interaction Analysis}
\begin{figure}[htbp]
    \centering
    \begin{flushright}
    \includegraphics[width=0.934\textwidth]{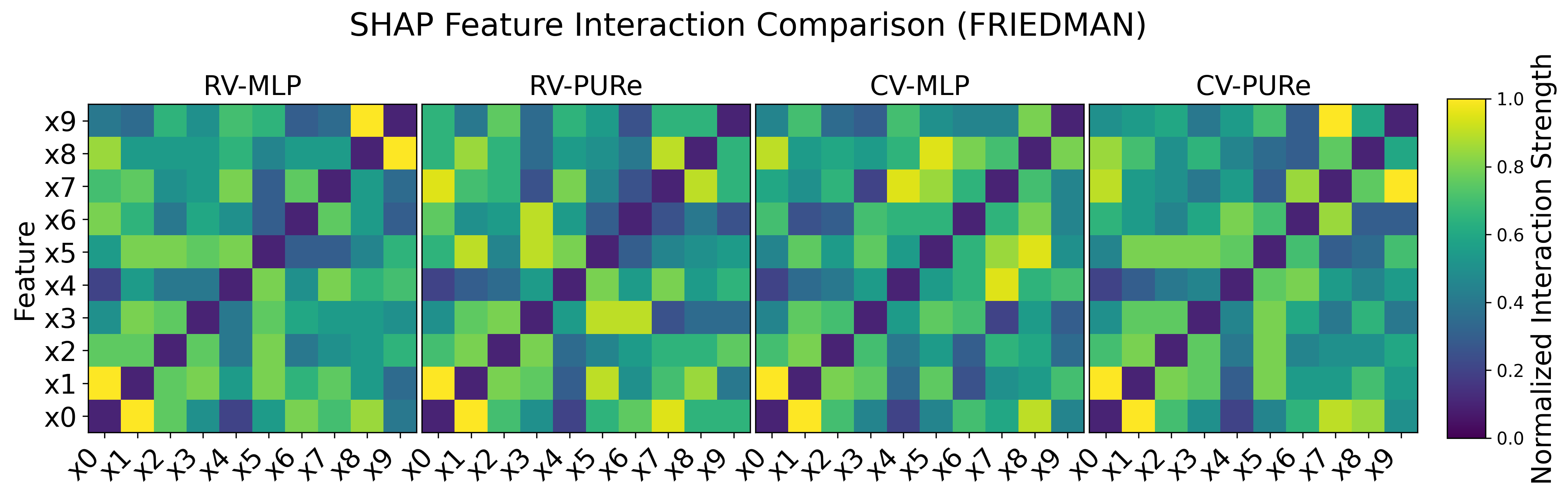}
    \end{flushright}
    \vspace{2em}
    \includegraphics[width=\textwidth]{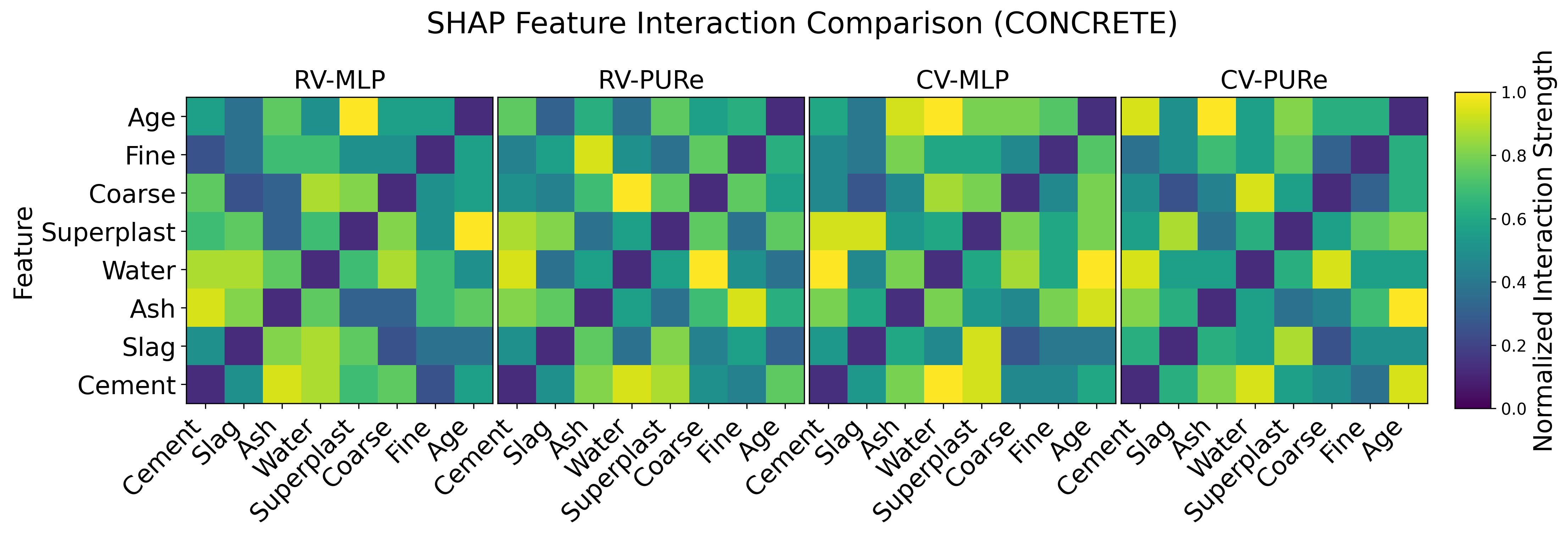}\\
    \vspace{2em}
    \includegraphics[width=\textwidth]{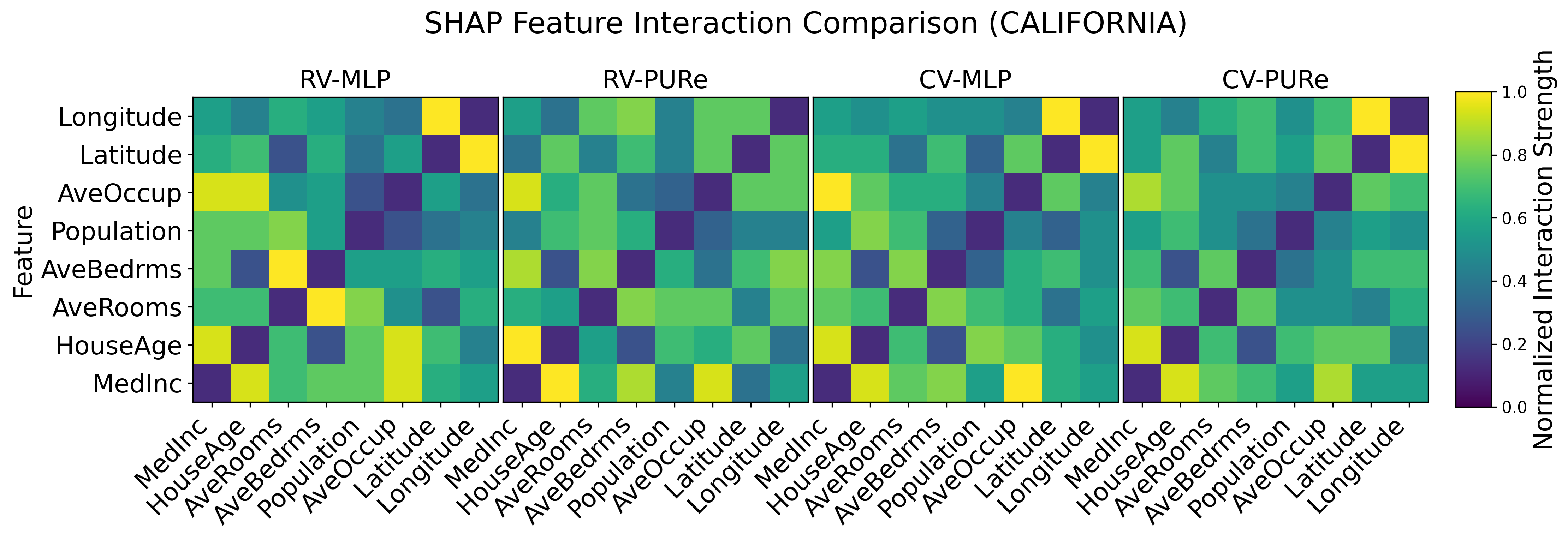}\\
    \vspace{2em}
    \caption{SHAP-based feature interaction comparisons on Friedman 1 (top), Concrete Compressive Strength (middle), and California Housing (bottom). For each dataset, interaction maps are shown for RV-MLP, RV-PURe, CV-MLP, and CV-PURe. Color intensity indicates normalized interaction strength.}
    \label{fig:shap_interactions}
\end{figure}

To evaluate the interpretability and structural alignment of the models, we analyze pairwise feature interaction patterns using SHAP-based, ranking-derived interaction maps. Figure~\ref{fig:shap_interactions} shows the normalized interaction maps for RV-MLP, RV-PURe, CV-MLP, and CV-PURe across the three benchmark datasets. These visualizations serve as a diagnostic of whether explicitly incorporating multiplicative units leads to interaction structures that are more concentrated and less diffuse than those produced by purely additive architectures.
\begin{table}[ht]
\centering
\caption{Quantifying the concentration of SHAP interaction maps (Fig.~\ref{fig:shap_interactions}). 
We report the normalized entropy $H_{\mathrm{norm}}$ (lower indicates more concentrated interactions) and the Herfindahl-Hirschman Index (HHI; higher indicates more concentrated interactions), both computed on the Top-$K$ off-diagonal interaction pairs (here $K{=}10$). Specifically, we normalize the Top-$K$ interaction strengths to a probability distribution $\{p_k\}_{k=1}^{K}$ and compute $H_{\mathrm{norm}} = -\sum_k p_k\log p_k / \log K$ and $\mathrm{HHI}=\sum_k p_k^2$. Best values per dataset are highlighted in bold (min for $H_{\mathrm{norm}}$, max for HHI). Metrics are computed on the same fixed-seed run as Fig.~\ref{fig:shap_interactions} and are intended as descriptive summaries rather than aggregated estimates.}
\label{tab:shap_concentration}
\setlength{\tabcolsep}{6pt}
\renewcommand{\arraystretch}{1.15}
\begin{tabular}{lcccc}
\toprule
Dataset & RV-MLP & RV-PURe & CV-MLP & CV-PURe \\
\midrule
Friedman  & 0.504 / 0.107 & 0.504 / 0.108 & 0.507 / 0.105 & \textbf{0.492 / 0.113} \\
Concrete  & 0.566 / 0.105 & 0.566 / 0.105 & 0.567 / 0.104 & \textbf{0.560 / 0.108} \\
California& 0.567 / 0.104 & 0.563 / 0.107 & 0.564 / 0.106 & \textbf{0.560 / 0.110} \\
\bottomrule
\end{tabular}
\end{table}

On the Friedman-1 dataset, where the target is generated from specific nonlinear interactions (most notably between $x_0$ and $x_1$), the interaction maps reveal clear differences in concentration. In Fig.~\ref{fig:shap_interactions} (top), the MLP baselines tend to distribute interaction strength across a broader set of feature pairs, beyond the dominant $(x_0,x_1)$ coupling. In contrast, product-unit models appear to display a more concentrated interaction landscape, with CV-PURe exhibiting the strongest concentration. This is corroborated by Table~\ref{tab:shap_concentration}: CV-PURe achieves the lowest normalized entropy ($H_{\text{norm}}$) and the highest HHI on Friedman, indicating that the Top-$K$ interaction signal is dominated by a small number of pairs, consistent with the ground-truth $(x_0,x_1)$ interaction.

A similar trend can be observed for the Concrete Compressive Strength dataset (middle panel). The interaction maps of the MLP baselines are comparatively dense, suggesting a more diffuse allocation of interaction strength across many pairs of mixture components. The PURe variants, especially CV-PURe, show more localized high-intensity regions indicating  a smaller subset of dominant material couplings (e.g., involving Cement, Water, and Age), while reducing diffuse background structure. Quantitatively, CV-PURe again achieves the most concise interaction profile, as shown in Table~\ref{tab:shap_concentration}.

For the California Housing dataset (bottom panel), interaction structure is inherently more heterogeneous, and all models highlight the key spatial dependency between Latitude and Longitude. However, PURe models, and in particular CV-PURe, produce interaction patterns that are more regular and less fragmented than those of the additive baselines. This is supported by the result shown in Table~\ref{tab:shap_concentration}, where CV-PURe yields the lowest $H_{\mathrm{norm}}$ and the highest HHI, indicating a more concise interaction map. Overall, across all three benchmarks, explicit multiplicative modeling leads to more structured and concentrated SHAP interaction patterns, and the complex-valued PURe variant shows the strongest effect.

\subsection{Sample Efficiency on Concrete}
\begin{figure}[htbp]
    \centering
    \includegraphics[width=0.95\linewidth]{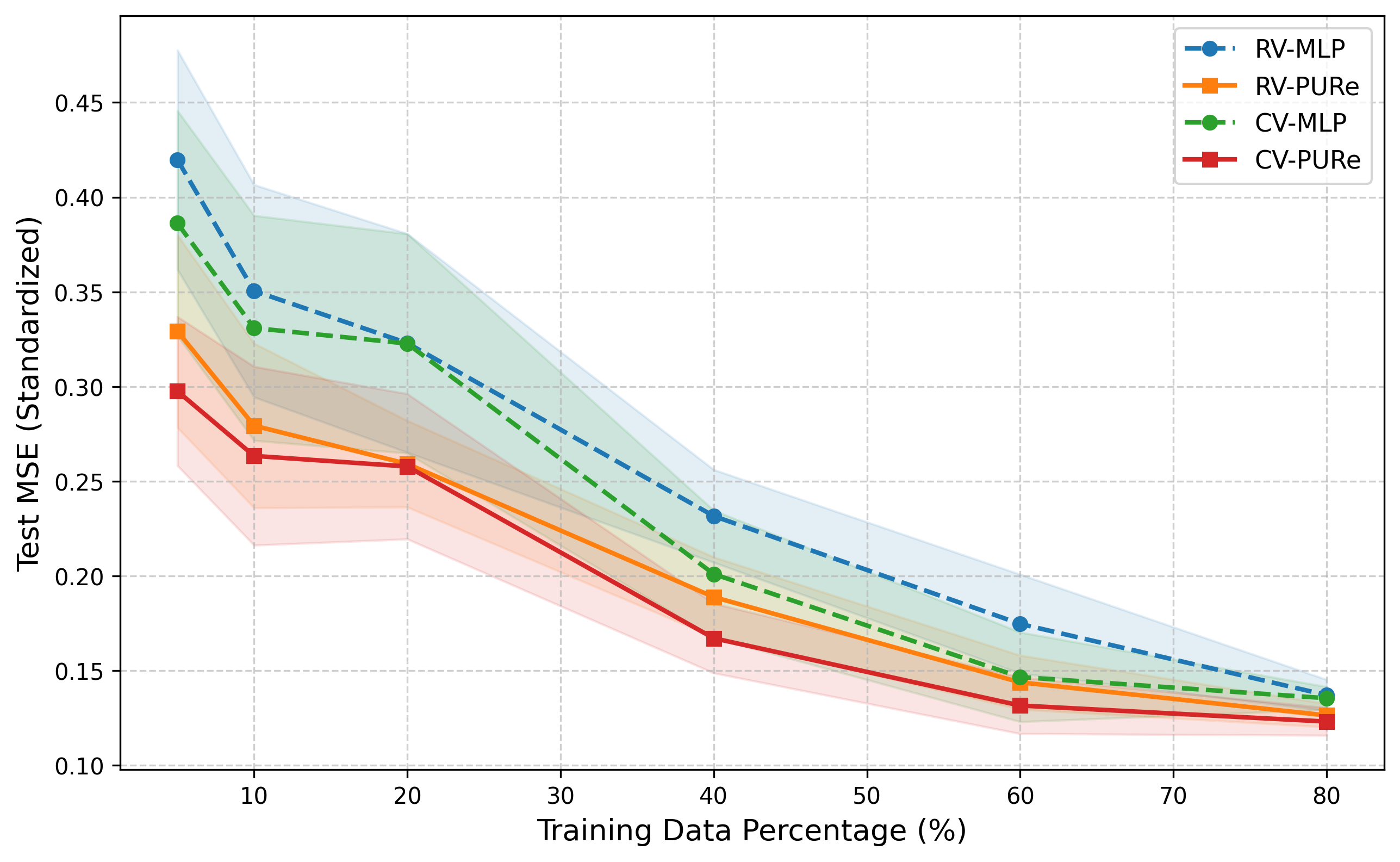}
    \caption{Sample efficiency comparison on the Concrete dataset.
    Test MSE (mean $\pm$ sample standard deviation over five runs) versus training data percentage.}
    \label{fig:sample_eff_concrete}
\end{figure}

To evaluate sample efficiency, we train all models on progressively smaller subsets of the Concrete dataset while keeping the validation and test sets fixed. Specifically, we used training fractions of \{5\%, 10\%, 20\%, 40\%, 60\%, 80\%\} and repeated each configuration over five random seeds. Figure~\ref{fig:sample_eff_concrete} shows the test MSE as a function of the available training data, and Table~\ref{tab:sample_eff_concrete} reports the corresponding numerical results (mean $\pm$ sample standard deviation).
\begin{table}[ht]
\centering
\setlength{\tabcolsep}{3pt}
\renewcommand{\arraystretch}{1.15}
\caption{Sample efficiency on Concrete: test MSE (mean $\pm$ sample standard deviation over five runs). Best results at each training fraction are highlighted in bold.}
\label{tab:sample_eff_concrete}
\begin{tabular}{lcccc}
\toprule
Train (\%) & RV-MLP & RV-PURe & CV-MLP & CV-PURe \\
\midrule
5  & 0.4197 $\pm$ 0.0579 & 0.3292 $\pm$ 0.0508 & 0.3862 $\pm$ 0.0594 & \textbf{0.2976} $\pm$ \textbf{0.0393} \\
10 & 0.3505 $\pm$ 0.0561 & 0.2794 $\pm$ \textbf{0.0434} & 0.3309 $\pm$ 0.0594 & \textbf{0.2634} $\pm$ 0.0472 \\
20 & 0.3230 $\pm$ 0.0577 & 0.2592 $\pm$ \textbf{0.0228} & 0.3227 $\pm$ 0.0579 & \textbf{0.2578} $\pm$ 0.0383 \\
40 & 0.2317 $\pm$ 0.0245 & 0.1889 $\pm$ 0.0211 & 0.2010 $\pm$ 0.0336 & \textbf{0.1672} $\pm$ \textbf{0.0184} \\
60 & 0.1747 $\pm$ 0.0261 & 0.1439 $\pm$ \textbf{0.0142} & 0.1467 $\pm$ 0.0236 & \textbf{0.1316} $\pm$ 0.0148 \\
80 & 0.1371 $\pm$ 0.0082 & 0.1264 $\pm$ \textbf{0.0062} & 0.1355 $\pm$ 0.0060 & \textbf{0.1232} $\pm$ 0.0073 \\
\bottomrule
\end{tabular}
\end{table}

Across the entire range of training data fractions, the product-unit models (RV-PURe and CV-PURe) consistently outperform the additive MLP baselines. This advantage is most prominent in the extremely low-data regime; at the 5\% training fraction, CV-PURe achieves a test MSE of 0.2976, which is significantly lower than the 0.4197 reported for RV-MLP. The visual gap between the PURe architectures (red and orange lines) and the MLP models (blue and green lines) in Figure~\ref{fig:sample_eff_concrete} confirms that explicit multiplicative modeling provides a much stronger inductive bias when supervision is severely limited.

A detailed inspection of Table~\ref{tab:sample_eff_concrete} reveals that CV-PURe attains the best overall performance at every training fraction. Furthermore, at the 5\% level, CV-PURe exhibits the smallest standard deviation ($\pm 0.0393$), indicating that the complex-valued parameterization not only improves accuracy, but also training stability when data is scarce. While the RV-PURe model also performs well and occasionally shows lower variability at mid-range fractions, CV-PURe remains the most efficient architecture, maintaining the lowest mean error throughout the experiment.

As the available training data increases from 5\% to 80\%, the performance gap between the architectures gradually narrows, as illustrated by the converging trajectories in Figure~\ref{fig:sample_eff_concrete}. However, even at the 80\% fraction, the product-unit models maintain a measurable lead, with CV-PURe reaching a final test MSE of 0.1232 compared to 0.1355 for CV-MLP. This indicates that while additive networks can eventually approximate complex nonlinear relationships with sufficient data, PURe models extract more information from fewer samples, demonstrating superior sample efficiency for modeling complex material dependencies.

\subsection{Ablation Study: Disentangling PU and Residual Roles}
To isolate the individual contributions of multiplicative product units and residual connections, we performed an ablation study on the Concrete dataset. We compare four variants: the standard MLP baseline; a Res-MLP, which incorporates residual connections into the MLP architecture in the same manner as the proposed PURe; a standalone product-unit network (PU) that removes all residual connections and adds activation functions between adjacent fully connected layers; and the PURe. Table~\ref{tab:ablation} reports the MSE for these configurations under both clean ($\sigma=0.00$) and noisy ($\sigma=0.05$) conditions.
\begin{table}[ht]
\centering
\setlength{\tabcolsep}{6pt}
\renewcommand{\arraystretch}{1.15}
\caption{Ablation study on the Concrete dataset. Test MSE (mean $\pm$ sample standard deviation over five runs).}
\label{tab:ablation}
\begin{tabular}{lcc}
\toprule
Model Variant & $\sigma = 0.00$ & $\sigma = 0.05$ \\
\midrule
MLP (baseline)           & 0.1118 $\pm$ 0.0041 & 0.1275 $\pm$ 0.0068 \\
Res-MLP (MLP + Residual) & 0.1209 $\pm$ 0.0038 & 0.1346 $\pm$ 0.0061 \\
PU (no residual)         & 0.1100 $\pm$ 0.0040 & 0.1269 $\pm$ 0.0064 \\
PURe (PU + Residual)     & \textbf{0.1095} $\pm$ \textbf{0.0015} & \textbf{0.1265} $\pm$ \textbf{0.0028} \\
\bottomrule
\end{tabular}
\end{table}

The results indicate that the explicit multiplicative bias provided by the product units is the primary driver of performance gains. As shown in Table~\ref{tab:ablation}, the PU variant without residual connections outperforms the baseline MLP, reducing the test MSE from 0.1118 to 0.1100 under clean inputs. In contrast, the Res-MLP variant actually results in a performance degradation compared to the baseline, with the error increasing to 0.1209. This suggests that for this specific regression task, adding residual connections to a standard additive architecture without the appropriate functional bias does not necessarily improve generalization.

The combination of both components in PURe yields the best overall results, achieving the lowest test MSE of 0.1095. Furthermore, the introduction of residual connections to the product-unit architecture significantly enhances training stability, as evidenced by the substantial reduction in standard deviation—from 0.0040 in the PU model to 0.0015 in PURe. This pattern holds true under noisy conditions ($\sigma=0.05$), where PURe maintains the highest accuracy and the lowest variability. These findings demonstrate that while multiplicative units provide the necessary inductive bias to capture complex feature interactions, the residual framework is essential for stabilizing the learning process and achieving optimal predictive performance.

\section{Analysis and Discussion}
\subsection{Interaction-aligned inductive bias}
The central empirical finding is that product-unit models are beneficial when the target is governed by nonlinear feature interactions. This confirms our intuition: product units directly parameterize multiplicative couplings, so the model does not need to synthesize interaction structure indirectly through many additive compositions. As a result, the learned dependency structure tends to be focused on a small set of informative cross-feature relations, rather than being spread across many weak and entangled interactions. This observation is quantitatively supported by a lower normalized interaction entropy and higher concentration (HHI) calculated from the Top-$K$ (here $K{=}10$) SHAP-based, ranking-derived interaction maps (Table~\ref{tab:shap_concentration}).

\subsection{Robustness under feature noise}
With Gaussian perturbations on the inputs, product-unit variants typically exhibit a smaller performance drop. A plausible explanation is that, once the model relies on a few stable, high-signal interactions, predictions become less sensitive to small, distributed perturbations across many weak interaction channels. In contrast, diffuse interaction patterns can accumulate noise effects, leading to larger error increases.

\subsection{Complex-valued modeling: expressiveness vs. variability}
Complex-valued product units often improve mean performance under a matched parameter budget, potentially because magnitude and phase offer a richer parameterization for representing nonlinear dependencies. The trade-off is a slightly increased run-to-run variability on some datasets, which is expected given the larger effective hypothesis space and higher sensitivity to initialization. Importantly, this variability manifests itself mainly around a stronger mean, rather than indicating systematic instability.

\subsection{Why residual connections matter here}
Residual pathways are particularly helpful for multiplicative blocks, where activations can be amplified or attenuated sharply and gradients may become less predictable. The skip connection provides a stable identity route that improves optimization reliability and reduces sensitivity to early training dynamics. This also explains why residualization is not guaranteed to help a plain MLP in shallow tabular settings, but becomes beneficial when paired with multiplicative transformations.

\subsection{Interpretability and sample efficiency}
Interaction analyses indicate that product-unit models yield more concentrated and structurally coherent pairwise dependency patterns, emphasizing a small set of dominant interactions while reducing diffuse low-strength couplings. This aligns with the sample-efficiency behavior: in low-data regimes, an interaction-aligned inductive bias effectively regularizes the hypothesis space, improving generalization when supervision is limited.

\section{Conclusion}
We introduced PURe, a product-unit residual architecture for tabular regression, and investigated both real- and complex-valued parameterizations under matched capacity. Across interaction-driven synthetic data and real-world benchmarks, product-unit models generally achieved lower prediction error than MLP baselines, with the strongest gains observed on the Friedman 1 and California Housing datasets. In particular, the complex-valued PURe variant reduced test MSE by up to about 35\% on the synthetic benchmark and by roughly 12--15\% under some noisy settings, while improvements on the Concrete dataset were more modest. In the low-data regime, the advantage of explicit multiplicative modeling became more pronounced, yielding about 10--30\% lower error depending on the training fraction. The learned interaction patterns also appeared more concentrated and structurally coherent in the SHAP-based analyses, suggesting that explicit multiplicative modeling provides a useful inductive bias for capturing nonlinear dependencies. Complex-valued PURe further improved mean performance in most settings, although it occasionally exhibited slightly higher variability across runs. Future work will focus on stronger quantitative validation of interaction structure, improved numerical stabilization for complex product units, and broader robustness evaluation under distribution shift and missing-feature settings.

\bibliographystyle{splncs04}
\bibliography{reference}

\end{document}